\documentclass[10pt,twocolumn,letterpaper]{article}

\usepackage[pagenumbers,algorithms,preprint]{wacv}

\usepackage{graphicx}
\usepackage{amsmath}
\usepackage{amssymb}
\usepackage{booktabs}
\usepackage{verbatim, algpseudocode, algorithm, algorithmicx, array, arydshln, pifont, capt-of, wrapfig, lipsum, makecell, tabulary, etoolbox, subcaption, fontawesome, xcolor}
\usepackage{multirow}

\usepackage[pagebackref,breaklinks,colorlinks]{hyperref}

\usepackage[capitalize]{cleveref}
\crefname{section}{Sec.}{Secs.}
\Crefname{section}{Section}{Sections}
\Crefname{table}{Table}{Tables}
\crefname{table}{Tab.}{Tabs.}

\def\wacvPaperID{1879}
\def\confName{WACV}
\def\confYear{2025}

\newcommand{\ModelName}{CerberusDet}

\usepackage{amsmath,amsfonts,bm}

\def\eqref#1{equation~\ref{#1}}

\def\1{\bm{1}}

\DeclareMathAlphabet{\mathsfit}{\encodingdefault}{\sfdefault}{m}{sl}
\SetMathAlphabet{\mathsfit}{bold}{\encodingdefault}{\sfdefault}{bx}{n}

\newcommand{\lr}{\alpha}

\newcommand{\Tau}{\scalebox{1.55}{$\tau$}}

\begin{document}
\definecolor{CornflowerBlue}{HTML}{0071BC}

\title{CerberusDet: Unified Multi-Dataset Object Detection}

\author{Irina Tolstykh \and Mikhail Chernyshov \and Maksim Kuprashevich\\
Layer Team, R\&D Department, SaluteDevices\\
{\tt\small irinakr4snova,imachernyshov,mvkuprashevich@gmail.com}
}
\maketitle

\begin{abstract}
\label{section:abstract}
Conventional object detection models are usually limited by the data on which they were trained and by the category logic they define. With the recent rise of Language-Visual Models, new methods have emerged that are not restricted to these fixed categories. Despite their flexibility, such Open Vocabulary detection models still fall short in accuracy compared to traditional models with fixed classes. At the same time, more accurate data-specific models face challenges when there is a need to extend classes or merge different datasets for training. The latter often cannot be combined due to different logics or conflicting class definitions, making it difficult to improve a model without compromising its performance. In this paper, we introduce \ModelName, a framework with a multi-headed model designed for handling multiple object detection tasks. Proposed model is built on the YOLO architecture and efficiently shares visual features from both backbone and neck components, while maintaining separate task heads. This approach allows \ModelName\ to perform very efficiently while still delivering optimal results. We evaluated the model on the PASCAL VOC dataset and Objects365 dataset to demonstrate its abilities. \ModelName\ achieved state-of-the-art results with 36\% less inference time. The more tasks are trained together, the more efficient the proposed model becomes compared to running individual models sequentially. The training and inference code, as well as the model, are available as open-source. \footnote{https://github.com/ai-forever/CerberusDet}
\end{abstract}
\begin{figure*}[ht]
\centering
\includegraphics[width=0.9\linewidth]{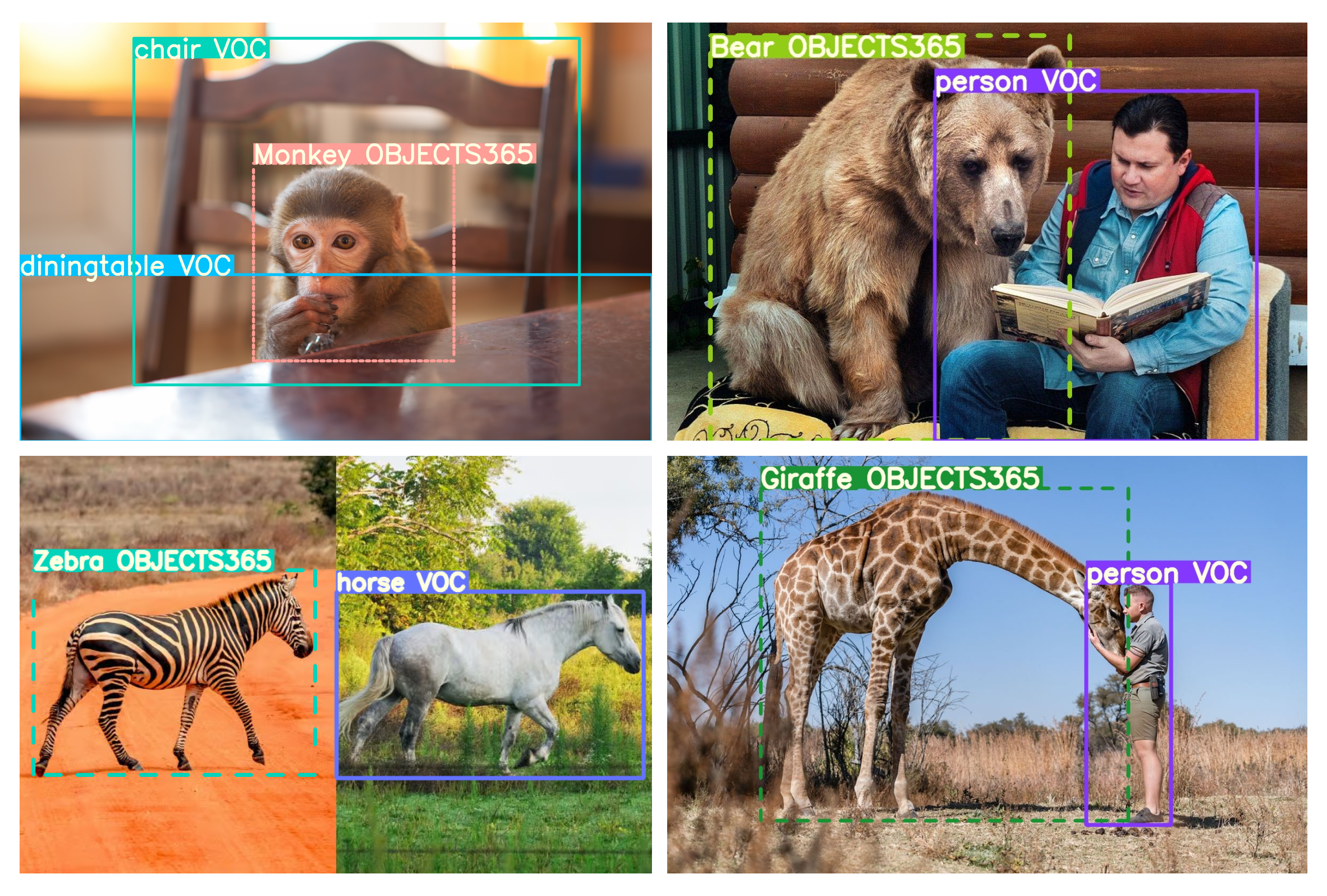}
\caption{Illustration of the work of the \ModelName\ trained on three datasets with different labels. We trained a model using the PASCAL VOC dataset and two subsets from the Objects365 dataset with animals and tableware categories. See training details in Section \ref{section:experiments}.}
\label{fig:example}
\end{figure*}

\section{Introduction}
\label{section:introduction}

Adding new categories to an existing real-time application that uses Object Detection (OD) involves several significant challenges. 
A key issue is that object categories annotated in one dataset might be unannotated in another, even if the objects themselves appear in images from the latter.
Additionally, merging different datasets often may be impossible because of differing annotation logic and incomplete class overlaps. 
At the same time, such applications require efficient pipelines, which limits the usage of separate data-specific models.

The goal of this work is \textit{to build a unified model trained on multiple datasets} that does not degrade in accuracy compared to individually trained models, while utilizing less computational budget. We present the \ModelName\ framework for training a single detection neural network on multiple datasets simultaneously. 
We also demonstrate an approach to identify the optimal model architecture, as not all tasks can be trained together. A notable challenge lies in determining which parameters to share across which tasks. Suboptimal grouping of tasks may cause negative transfer \cite{standley2020tasks}, the problem of sharing information between unrelated tasks. 
Additionally, with computational resource constraints, proposed approach allows us to choose the architecture that fits the requirements. To evaluate the proposed \ModelName\ model, we conduct experiments with open data and obtain results comparable to separated data-specific state-of-the-art models, but with one unified neural network. The presented architecture is based on YOLO\cite{yolov8}."

An alternative approach to extending the detector model with new categories is the use of Open-Vocabulary Object Detectors (OVDs)\cite{gu2021open}, which have recently gained popularity. However, OVDs often lack the accuracy of data-specific detectors, require a lot of training data, and are prone to overfitting to base classes \cite{jianzong2024survey,chaoyang2024survey}. We prioritize high accuracy over the flexibility of OVDs. The proposed architecture allows us to add new classes as needed while preserving the accuracy of previously learned ones, making our approach more suitable for the required needs. Notably, this approach has been deployed and validated in our production environment, demonstrating its robustness and reliability in practical applications.

The key contributions of our paper are as follows:
\begin{itemize}
\item We provide a study of various methods for multi-dataset and multi-task detection, exploring different parameter-sharing strategies and training procedures.
\item We present several experimental results using open datasets, offering insights into the effectiveness of various approaches.
\item We introduce a novel framework that can be tailored to different computational requirements and tasks, featuring a multi-branch object detection model named \ModelName.
\item We publicly release the training and inference code, along with the trained model, to encourage further research and development in this area.
\end{itemize}
\section{Related Works} 
\label{section:related_work}

\textbf{Object Detection} There are many different detection models. Two-stage detectors, such as Faster R-CNN \cite{ren2015faster} and Cascade R-CNN \cite{cai2018cascade} first generate region proposals, which are then refined and classified. Single-shot convolutional detectors like YOLO \cite{redmon2016you}, SSD \cite{liu2016ssd} or EfficientDet \cite{tan2020efficientdet} skips the region proposal stage and produce final localization and labels prediction at once.
Recently popularized detection transformers like DAB-DETR \cite{liu2022dab} or CO-DETR \cite{zong2023detrs} uses a transformer encoder-decoder architecture to predict all objects at once.

We built \ModelName\ model based on the implementation of the YOLO architecture by Ultralytics \cite{yolov8}, as YOLOv5/YOLOv8 models are fast and achieve SOTA results on various tasks. YOLOv5 uses anchors for the detection head which is composed of convolutional layers for multi-scale features. YOLOv8 is an anchor-free model with a decoupled head to process separately objectness, classification, and bounding box regression tasks based on multi-scale features. 

\textbf{Multi-Task Learning} (MTL) aims to improve both efficiency and prediction accuracy for individual tasks over separately trained models. The two most commonly used ways to perform MTL are hard or soft parameter sharing of hidden layers \cite{overview}. Authors of \cite{eigen2015predicting, liao2016understand, nekrasov2019real, doersch2017multi, long2017learning, leang2020dynamic, chennupati2019multinet++} apply the first one to share most of the parameters between all tasks and to find a representation that captures all of the tasks. Soft parameter sharing is utilized in works \cite{duong2015low, yang2016trace, duong2015low, misra2016cross, yang2016trace}, where individual tasks possess their own parameter sets interconnected either through information sharing or by requiring parameter similarity. 


Most multi-task models in the computer vision domain focus on addressing different CV tasks, such as classification and semantic segmentation \cite{liao2016understand}; segmentation, depth and surface normals \cite{eigen2015predicting}. UberNet \cite{kokkinos2017ubernet} learns 7 computer vision problems under a single architecture. GrokNet \cite{bell2020groknet} learns unified representation to solve several image retrieval and a large number of classification tasks. In this paper, we address to MTL to tackle various detection tasks.

To design the optimal multi-task network architecture the authors of \cite{standley2020tasks, branched, sun2020adashare, bruggemann2020automated, zamir2018taskonomy, fifty2021efficiently} apply different strategies based on understanding task relationships. In this work we use representation similarity analysis (RSA) \cite{rsa,dds,cka} method to estimate the task affinity similar to \cite{branched}.

Different optimization techniques were proposed in \cite{chen2018gradnorm, ica, mgda, yu2020gradient} for MTL systems, which aim to reduce conflicts between tasks by adjusting the direction of task gradients. In this paper, we employ a gradient averaging method, but any other optimization method can be utilized  as well for training the proposed \ModelName\ model.

\textbf{Multi-Dataset Object Detection} aims to leverage multiple datasets to train one visual recognition model to detect objects from different label spaces. Some works \cite{wang2019towards, xu2020universal} build specific modules to adapt feature representations related to different domains. Others \cite{zhou2022simple, zhao2020object} train one model with unified multi-dataset label spaces. To create a detector with an unified label space across all datasets, the authors of \cite{zhou2022simple} automatically learn mappings between the common label space and dataset-specific labels during training. The current paper focuses on a model with shared parameters but dataset-specific outputs. The authors of \cite{zhao2020object} train a detection model with pseudo ground truth for each dataset generated by task-specific models to merge label spaces, while our framework does not require annotations from different datasets to be combined. ScaleDet \cite{chen2023scaledet} also unifies label space from multiple datasets by utilizing text CLIP \cite{clip} embeddings.

\textbf{Open-Vocabulary Object Detection} (OVD) models aim to recognize objects of categories not present at training time. Novel categories are described by text inputs and the detectors are try to establish a semantic connection between object regions and object labels chosen from a possibly very large vocabulary \cite{gu2021open, feng2022promptdet, bianchi2023devil}. The association of objects and labels typically is done through large pre-trained vision-language matching methods like CLIP \cite{cheng2401yolo, gu2021open, feng2022promptdet, kuo2022f}. 

OVD models may be used for expanding the label set of a detector model if pretrained models have knowledge of the target data domain, aligning textual embedding space with visual features during training \cite{feng2022promptdet,kuo2022f,minderer2022simple, ma2022rethinking, cheng2401yolo}. The authors of \cite{gu2021open} train an open-vocabulary detector based on CLIP text and image embeddings to detect 1,203 categories from the LVIS dataset. They initially train the detector on base categories and then expand it to cover all rare categories in the dataset.

\section{Model} 
\label{section:method}

\subsection{Method}

\begin{figure*}
\centering
\includegraphics[width=0.9\linewidth]{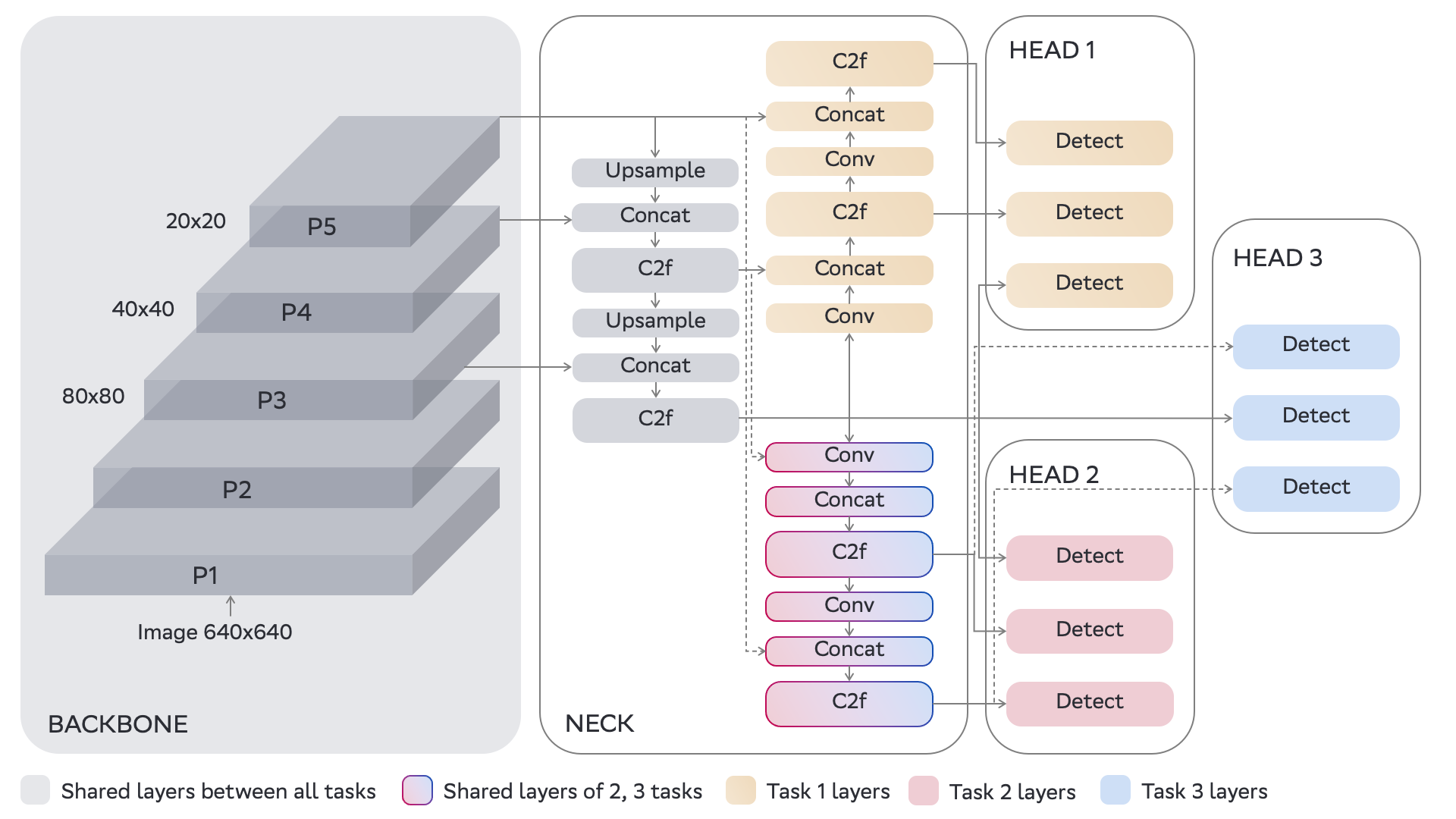}
\caption{Diagram of an example of the \ModelName\ architecture based on YOLOv8, illustrated with three tasks. Each neck module can be shared between tasks or be task-specific. The \ModelName\ model optimizes computational resources by sharing all backbone parameters across tasks, while each task retains its own unique set of parameters for the head.}
\label{fig:scheme}
\end{figure*}

\begin{figure}[t]
\centering
\includegraphics[width=8.2cm]{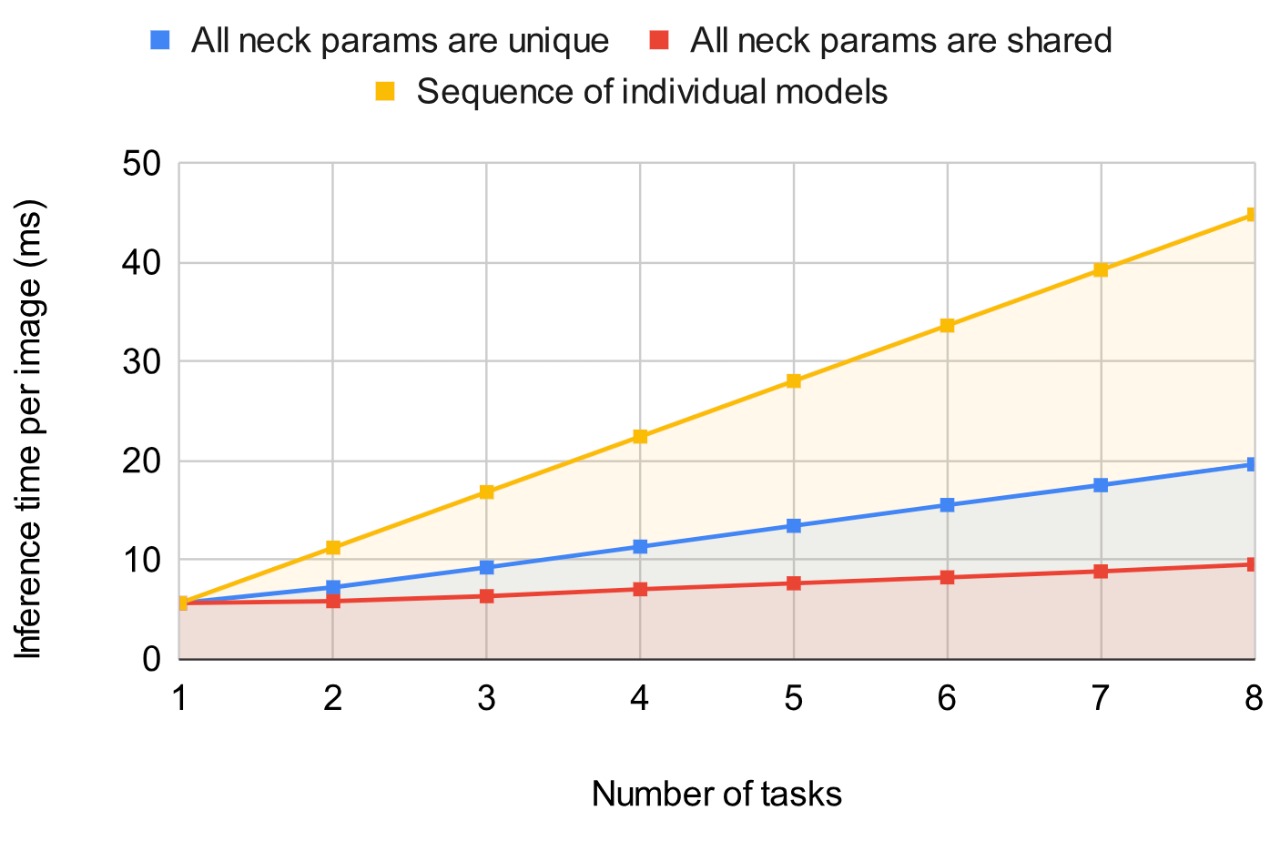}
\caption{Comparison of inference time of YOLOv8x-based \ModelName\ models and the sequence of individual models. Measurements were
made with FP16 precision on a V100 GPU with a batch size of 32.}
\label{fig:inference_time}
\end{figure}

In this paper, we propose the \ModelName\ model that allows multiple detection tasks to be learned in a shared model. Each detection task is a separate task, which employs its own dataset with the unique set of labels.

The \ModelName\ model is built upon the YOLO \cite{yolov8} architecture. It optimizes computational resources by sharing all backbone parameters across tasks, while each task retains its own unique set of parameters for the head. The neck layers can either be shared or specific to  the task. One of the possible variants of YOLOv8-based \ModelName\ architecture for three tasks is illustrated in Figure \ref{fig:scheme}. With the standard YOLOv8x architecture and 640 input image resolution, the model backbone consists of 184 layers and 30M parameters. The neck has 6 shareable modules with 134 layers and 28M parameters. Each head consists of 54 layers and 8M parameters.

By sharing the backbone across multiple tasks, our training approach achieves significant computational budget economy compared to the sequential inference of separate models for each task. Figure \ref{fig:inference_time} illustrates the inference speed of \ModelName, which is based on the YOLOv8x architecture. The figure compares the inference times for two scenarios: one where all neck parameters are task-dependent, and another where these parameters are shared across tasks. The results highlight the computational efficiency gained through parameter sharing.

\subsection{Parameters sharing}

We decided to employ the hard parameter sharing technique for multi-task learning, given its demonstrated efficiency and its ability to enhance per-task prediction quality by leveraging information across tasks during training \cite{overview}. Hard parameters sharing allows us to have sets of parameters that are shared across tasks, and sets of parameters that are task-specific. Based on YOLO architecture we have sets of sharable parameters at the module level. E.g. YOLOv8x have 6 parameterized neck modules, so each task may share each of them with another task.

To decide what modules to share across which tasks we employ the Representation Similarity Analysis \cite{rsa,dds,branched} method to estimate task similarity at each neck module that can be shared or task-specific. Then for each possible architecture variant we calculate an RSA-based similarity score ($\mathit{rsa\ score}$) and $\mathit{computational\ score}$. The first one shows the potential performance of an architecture and the second one evaluates its computational efficiency. Within the available computational budget, we select the architecture with the best $\mathit{rsa\ score}$. Let the architecture contain $l$ shareable modules and we have $N$ tasks, the algorithm for selecting the architecture looks as follows:
\begin{enumerate}
    \item[-] Select a small representative subset of images from the test set of each task.
    \item[-] Using task-specific models, extract features for the selected images from each module.
    \item[-] Based on the extracted features, calculate Duality Diagram Similarity (DDS) \cite{dds} - computing pairwise (dis)similarity for each pair of selected images. Each element of the matrix is the value of (1 - Pearson’s correlation).
    \item[-] Using the Centered Kernel Alignment (CKA) \cite{cka} method on the DDS matrices, compute representation dissimilarity matrices (RDMs) - an $NxN$ matrix for each module. Each element of the matrix indicates the similarity coefficient between two tasks.
    \item[-] For each possible architecture, using values from the RDM matrices, compute the $\mathit{rsa\ score}$. It is calculated as the sum of the task dissimilarity scores at every location in the shareable model layers. It is defined as $\mathit{rsa\ score} = \sum_{m=1}^{l} S_m$, where $S_m$ (equation \ref{eq:rsa}) is found by averaging the maximum distance between the dissimilarity scores of the shared tasks in the module l.
    \item[-] For each possible architecture calculate $\mathit{computational\ score}$ using formula \ref{eq:comp_score}.
    \item[-] We select the architecture with the best combination of $\mathit{rsa\ score}$ and $\mathit{computational\ score}$ (the lower is the better), or we choose the architecture with the lowest $\mathit{rsa\ score}$ within the set constraint on $\mathit{computational\ score}$.
\end{enumerate}

\begin{align}
\label{eq:rsa}
    S_m &= \frac{1}{|\{\Tau_i, \ldots, \Tau_k\}|} \times \\
    &\sum_{j=i}^{k} \max \left\{ RDM(j, i), \ldots, RDM(j, k) \right\} \notag
\end{align}
where $\{\Tau_i, \ldots, \Tau_k\}$ - shared tasks at module $l$.

\begin{equation}
\label{eq:comp_score}
   \mathit{computational\ score} = \frac{\mathit{inference\_time}}{(N * \mathit{single\_inference\_time})}
\end{equation}

\begin{figure}[t]
\centering
\includegraphics[width=8.2cm]{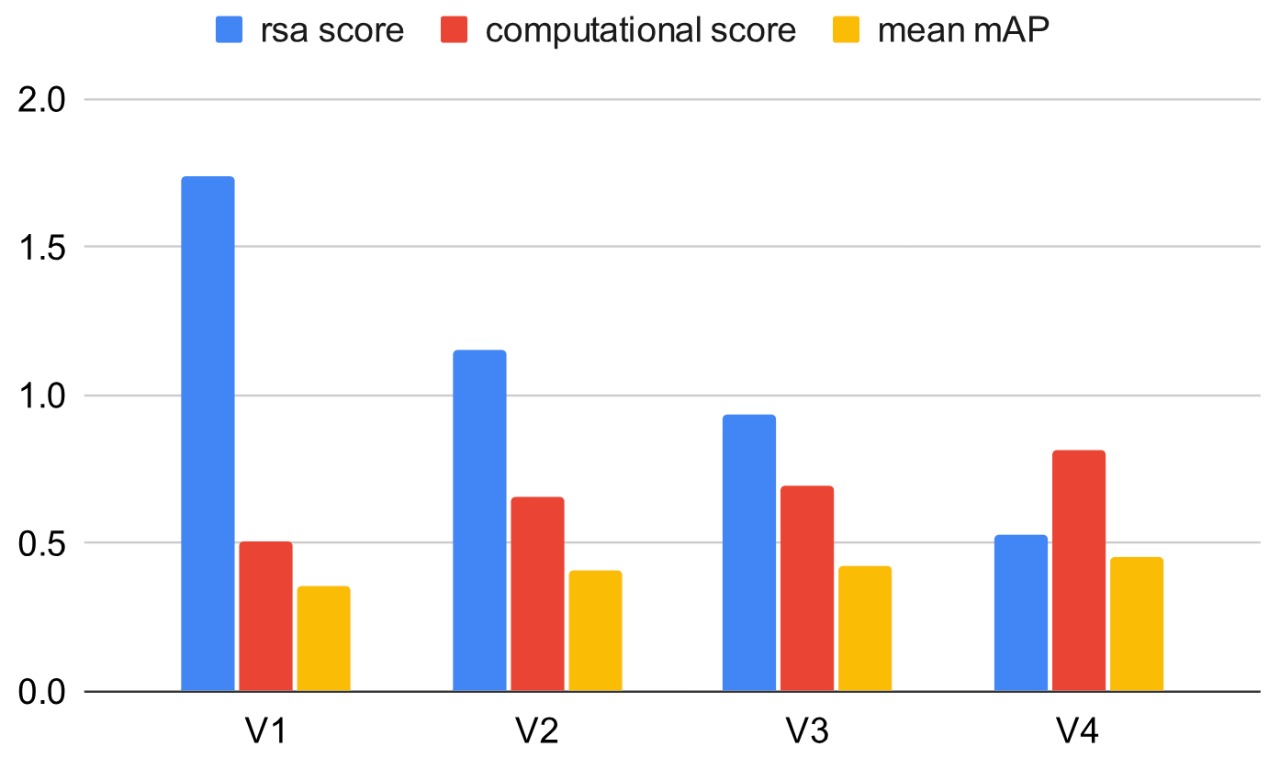}
\caption{$rsa\ score$, $computational\ score$ and average mAP of 4 different models trained for 3 tasks.}
\label{fig:rsa}
\end{figure}

To evaluate the chosen approach, we selected 4 architectures with different RSA scores and computational scores, trained models and compared the average metric values. Figure \ref{fig:rsa} demonstrates that model accuracy increases along with RSA score decreases and computational complexity increases. To calculate the computational score, the V100 GPU was used and the batch size was equal to 1.

\subsection{Training procedure}
\label{subsection:training}

Let's consider a set of tasks $\{\mathit{\Tau_1, \ldots, \Tau_n}\}$. Different combinations of these tasks may share a set of model parameters. Let $\theta_{shared} = \{\theta_{i..k}, \ldots \theta_{j..m}\}$ be the sets of shared parameters between different groups of tasks  $\{i, \ldots,k\}, \ldots, \{j, \ldots,m\}$. Algorithm \ref{algo:train} represents the end-to-end learning process of the proposed \ModelName\ model.
During training, we iterate through tasks, sample mini-batches from the corresponding dataset, calculate the loss and gradients for the parameters related to the current task. Next, we average the gradients for shared parameters of each task group and update their values according to equation \ref{eq:average}.

\begin{equation}
\label{eq:average}
    \theta_{\{i,\ldots,k\}} = \theta_{\{i,\ldots,k\}} - (\lr * \frac{1}{|\{i,\ldots,k\}|} * \sum_{j \in \{i,\ldots,k\}} \dfrac{\partial L_j}{\partial \theta_{\{i,\ldots,k\}}})
\end{equation}

where $\{i,\ldots,k\}$ represents the group of tasks with shared parameters $\theta_{\{i,\ldots,k\}}$, \(\lr\) is the learning rate, \(L_j\) is the loss for task \(j\).

\begin{algorithm}
\caption{\ModelName\ training process}
\label{algo:train}
\begin{algorithmic}[1] \small
\Require{
    \Statex $N$ -- total number for tasks,
  \Statex $\{\mathit{D_1,\ldots, D_n}\}$ -- datasets for tasks $\{\mathit{\Tau_1, \ldots, \Tau_n}\}$,
  \Statex $\{\mathit{\theta_1, \ldots, \theta_n, \theta_{shared}}\}$ -- task-specific and shared parameters,
  \Statex $\{\mathit{L_1, \ldots, L_n}\}$ -- task-specific loss functions
}
\While{training not converged}
  \State $g_{shared} \gets 0$
  \For{$i = 1$ to $N$}
    \State $g_i \gets 0$
    \State $X_i \gets \text{a mini batch from } \mathit{D_i} \text{ for task } \mathit{\Tau_i}$
    \State \textcolor{CornflowerBlue}{// Compute task-specific model output}
    \State $Y \gets \mathit{M}(X_i, \Tau_i)$
    \State \textcolor{CornflowerBlue}{// Calculate gradients for task-specific parameters}
    \State $g_{i} \gets \dfrac{\partial L_i}{\partial \theta_i}(Y, Y^*)$
    \State \textcolor{CornflowerBlue}{// Accumulate gradients for shared parameters}
    \State $g_{shared} \gets g_{shared} + \dfrac{\partial L_i}{\partial \theta_{shared}}(Y, Y^*)$
  \EndFor
  \State \textcolor{CornflowerBlue}{// Update parameters}
  \For{$i = 1$ to $N$}
    \State $\theta_{i} \gets \theta_{i} - \lr * g_i $
  \EndFor
  \State $\theta_{shared} \gets \theta_{shared} - \text{learning rate} \cdot \text{avg}(g_{shared}, \mathit{M})$  \Comment{Eq. \ref{eq:average}}
\EndWhile
\end{algorithmic}
\end{algorithm}

The speed and effectiveness of joint training are strongly influenced by the loss functions of individual tasks. Since these loss functions can have different natures and scales, it is essential to weigh them correctly. To find the optimal weights for the loss functions, as well as other training hyperparameters, we employ the hyperparameter evolution method.

During the training process, we discovered that the model's performance significantly suffers if the samples within each batch are not balanced carefully and thoroughly. To address this, we implemented a strategy to ensure that all classes are adequately represented in each iteration according to their frequency in the dataset.

\subsection{The impact of training settings}

The described in the previous two sections techniques were used in a series of experiments with our proprietary data. Table \ref{table:proprietary_results} presents the results of the impact of each technique. We use proprietary data in these experiments, as they exhibit a sufficient level of inter-task consistency to ensure the clarity of the experiments. Models were trained for 3 tasks, where the baseline being an architecture where all parameters of the model, except for the heads, were shared among tasks. The dataset of the first task comprises 22 categories with 27,146 images in the training set and 3,017 images in the validation set. The dataset of the second task consists of 18 categories with 22,365 images in the training set and 681 images in the validation set. The dataset of the third task comprises 16 categories with 17,012 images in the training set and 3,830 images in the validation set. To compare the influence of architecture search method on the result, we also trained the model, where all neck parameters are task-specific. Then, we compare the accuracy improvement of the discovered architecture relative to it.

Mentioned models were built on top of YOLOv5x with an input image resolution of 640x640, measurements were made with FP16 precision on the V100 GPU.

\begin{table*}[t]
\centering
\renewcommand{\arraystretch}{1.3}
\begin{tabular}{>{\raggedright\arraybackslash}m{8.8cm}|>{\centering\arraybackslash}m{1cm}}
\hline
 \multicolumn{1}{c|}{Methods applied during training} & \multicolumn{1}{c}{$\varDelta$ mAP@0.5:0.95} \\ [1ex]
\hline
\hline
 w/o balanced sampling, w/o arch search, w/o hypersearch  & +0\% \\[0.5ex] 
 \hline
 with balanced sampling & +5\% \\[0.5ex] 
 \hline
 with balanced sampling, all neck+head layers are task specific & +11\% \\[0.5ex] 
 \hline
 with balanced sampling, with arch search & +1\% \\[0.5ex] 
 \hline
 with balanced sampling, with arch search, with hypersearch & +1.2\% \\[0.5ex]
\hline
\end{tabular}
\caption{The impact of sequentially applying various techniques in YOLOv5x-based \ModelName\ training for 3 tasks. The baseline model shares all parameters between tasks, except for the heads. It was trained without balanced sampling and without hyperparameters tuning.}
\label{table:proprietary_results}
\end{table*}

\section{Open-source datasets experiments
} 
\label{section:experiments}

Within this section, we outline \ModelName's experimental setup, results, and training configuration. Additionally, we perform a comparative analysis that includes \ModelName\ and standalone YOLOv8 models using public datasets. 

Notably, our comparison also incorporates the Open Vocabulary Detector YOLO-Worldv2-X \cite{cheng2024yoloworld}, which was trained on a comprehensive set of base classes from the Objects365, GoldG \cite{kamath2021mdetr}, and CC3M \cite{sharma2018conceptual} datasets at a resolution of 640. This inclusion enables us to evaluate the performance of a model employing a zero-shot approach for class expansion and comparison with traditional detection models.

Our analysis includes the presentation of metrics such as mean average precision (mAP): mAP@0.5 and mAP@0.5:0.95, alongside measurements of the models inference speed with FP16 precision on a V100 GPU.

\subsection{Datasets}
\label{subsection:datasets}
We conducted experiments on two publicly available datasets: PASCAL VOC \cite{pascal-voc-2012} and Objects365 \cite{9009553}. We conducted two experiments. The first was done with the original datasets, and the second used the PASCAL VOC dataset and two Objects365 subsets to simulate training on 3 datasets with different domains. The first subset was selected to include only images containing specific animal categories not present in the PASCAL VOC dataset. The second subset contains objects from various tableware categories.

In both cases, standard datasets with predefined splits are used. For Objects365, these include the train and val sets, while for PASCAL VOC, they encompass train/val/test for the year 2007 and train/val for the year 2012. 
To create an Objects365 subset, all non-empty images with annotations containing at least one object belonging to the specific class were used. Empty images without annotations or images with objects from unrelated classes were filtered out. A detailed list of classes and the filtering code are available in the project repository.

The PASCAL VOC dataset comprises 20 classes with 16,551 images in the training set and 4,952 images in the validation set. For training, we combined the trainval sets of PASCAL VOC 2007 and PASCAL VOC 2012, and validated on the test set of PASCAL VOC 2007. The original Objects365 dataset consists of objects from 365 classes, with 1.742M images in the training set and 79,578 images in the validation set.
The Objects365 animals and Objects365 tableware subsets contain 19 and 12 classes, respectively. The training sets include 14,295 and 269,675 images respectively, while the validation sets include 5,413 and 24,645 images, respectively.

\begin{table*}[t]
\centering
\renewcommand{\arraystretch}{2}
\begin{tabular}{c|c|c|c|c|c}
\hline
Model & Train Set & Test Set & mAP 0.5 & mAP 0.5:0.95 & Speed(ms) \\ \hline \hline
YOLOv8            & \makecell{VOC 2007/VOC 2012}       & VOC 2007                 &    $0.92$     &   $0.76$      & 5.6  \\ \hline
YOLOv8    & \makecell{O365 }        & \makecell{O365}          & $0.38$ &  $0.29$     & 5.6  \\ \hline
YOLOv8    & \makecell{O365 animals }        & \makecell{O365 animals}          & $0.55$ &  $0.43$     & 5.6  \\ \hline
YOLOv8    & \makecell{O365 tableware }        & \makecell{O365 tableware}          & $0.68$ &  $0.56$     & 5.6  \\ \hline
\multirow{4}{*}{YOLO-Worldv2-X$^\dagger$} & \multirow{4}{*}{\makecell{O365 + GoldG + CC3M-Lite}} & VOC 2007 & 0.83 & 0.69 & \multirow{4}{*}{7.7} 
\\ \cline{3-5}
                       &                            & \makecell{O365} & 0.31 & 0.24 &
\\ \cline{3-5}
                       &                            & \makecell{O365 animals} & 0.37 & 0.3 & \\ \cline{3-5}
                       &                            & \makecell{O365 tableware} & 0.50 & 0.40 &                          
                       \\ \hline
\multirow{2}{*}{\ModelName} & \multirow{2}{*}{\makecell{VOC 2007/VOC 2012 + O365 }} & VOC 2007                 & $0.93$ & $0.77$ & \multirow{2}{*}{7.2} \\ \cline{3-5}
                       &                             & \makecell{O365 }          &   $0.46$ & $0.36$ &
                       \\ \hline
\multirow{2}{*}{\ModelName} & \multirow{2}{*}{\makecell{VOC 2007/VOC 2012 + O365 animals }} & VOC 2007                 & $0.92$ & $0.75$ & \multirow{2}{*}{7.2} \\ \cline{3-5}
                       &                             & \makecell{O365 animals }          &   $0.57$ & $0.43$ &
                       \\ \hline
\multirow{3}{*}{\ModelName} & \multirow{3}{*}{\makecell{VOC 2007/VOC 2012 + \\ O365 animals + \\ O365 tableware }} & VOC 2007                 & $0.93$ & $0.76$ & \multirow{3}{*}{10} \\ \cline{3-5}
                       &                             & \makecell{O365 animals }          &   $0.54$ & $0.42$ &
                       \\ \cline{3-5}
                       &                             & \makecell{O365 tableware }          &   $0.68$ & $0.56$ &
                       \\ \hline
\end{tabular}
\caption{Comparison of the performance of \ModelName\ models trained on two and three datasets, the OVD detector, and YOLOv8 dataset-specific models. Inference time measurements were conducted on a V100 GPU with a batch size of 32 and FP16 precision. $^\dagger$ indicates that the Ultralytics version of the model was used.}
\label{table:results}
\end{table*}

\subsection{Models Training}

Initially, we trained 4 individual YOLOv8x models on each of the datasets mentioned in section \ref{subsection:datasets}. These models were trained with a 640x640 input resolution, a batch size of 32, and mixed precision, serving as a baseline.

Then we trained the \ModelName\ model on two datasets, Objects365 and PASCAL VOC, on datasets whose sizes differ by a factor of 100.
We slightly modified the training procedure described in section \ref{subsection:training} so that during each epoch, the model saw each dataset only once. To achieve this, we skipped batches from the smaller VOC dataset. Thus, in each iteration, the model saw samples from the Objects365 dataset, while samples from the VOC dataset were fed into the model only every 104 iteration. The batch size was set to 40. The model converged in 38 epochs.

We also trained the \ModelName\ model on three datasets: Objects365 tableware, Objects365 animals, and PASCAL VOC. Similarly, there was a significant imbalance in the amount of data between the first and the other datasets. For training this model, we also modified the training process so that in each iteration, we used a different number of training examples from each dataset: 40, 4, and 4, respectively. The model was converged in 29 epochs.

We also trained the \ModelName\ model on two datasets: PASCAL VOC and Objects365 animals to assess how the metrics change when a new dataset is added. We used standard training procedure, batch size of 32 and one V100 GPU to train this model. 

We used the same set of hyperparameters for all training procedures, inherited from YOLOv8x single models training. \ModelName\ models leverage the YOLOv8x architecture as the backbone. The vanilla YOLOv8 head is used for each detection task, with all neck parameters being dataset-specific. Additionally, we utilized transfer learning by initializing \ModelName's backbone and neck with a pre-trained model on COCO povided by the YOLOv8 authors, while the heads parameters were initialized randomly. 

\ModelName\ models were trained with mixed precision, the SGD optimizer and synchronized batch normalization using 8 H100 GPUs. The input resolution was set to 640x640 pixels.

\subsection{Experimental Results}

The experimental results are presented in Table \ref{table:results}.

We can observe that all three \ModelName\ models either achieve the same accuracy as the individual YOLOv8 models or outperform them. When comparing the \ModelName\ trained on two datasets (VOC, O365 animals) with the one trained on three (VOC, O365 animals, O365 tableware), we can see that adding another dataset to the training has minimal impact on the accuracy for the first two datasets.

In comparison with YOLO-Worldv2-X, a specialized \ModelName\ models demonstrate a significant advantages. This comparison confirms the hypothesis that OVD detectors, while flexible and capable of rapidly adapting to new classes, generally fall short of the accuracy achieved by specialized detectors like \ModelName.

When evaluating model inference speed, \ModelName\ remains efficient, performing inference for two datasets in 7.2 milliseconds on a single NVIDIA V100 GPU with FP16 precision, faster than separate YOLOv8 models (11.2 ms). Extending \ModelName\ to three datasets increases inference time to 10 milliseconds, which is still one-third faster than using three separate models. For two tasks, \ModelName\ also shows a slight advantage over YOLO-Worldv2-X, with 7.2 ms compared to 7.7 ms.

The results demonstrate effectiveness of the \ModelName\ across diverse datasets, even when a significant imbalance in the amount of data is present. Results also highlights an advantage over the zero-shot approach while maintaining flexibility in class expansion. 
\section{Limitations}
\label{section:limitations}

The training process outlined in the paper is highly sensitive to optimization hyperparameters such as learning rate, loss weights, momentum, and weight decay. Therefore, we recommend conducting a hyperparameter search to achieve the best training results. The necessary scripts for this process are provided with the code.

Additionally, in certain cases, it may be beneficial to consider different multi-gradient descent algorithms instead of gradient averaging. Notable examples include MGDA\cite{mgda} and Aligned-MTL\cite{ica}.

When training a model on multiple tasks, it is important to identify which tasks can share more parameters and which require fewer shared parameters. To address this, we use the RSA algorithm, which, while requiring models trained for individual tasks, is still faster than iteratively testing all possible architecture variations with different parameter-sharing schemes. However, if training such models is not feasible, our approach can still be applied using a parameter-sharing scheme where all neck layers are task-dependent. While this may not be the most optimal solution, it remains more efficient than sequentially inferring single models.

\section{Conclusions}
\label{section:conclusions}

In this work, we introduced \ModelName, a scalable and adaptive framework for multi-task object detection. The proposed method achieves comparable to separated data-specific state-of-the-art models while utilizing approximately 36\% less computational budget in case of training for two tasks. The more tasks are trained together, the more efficient the proposed model becomes compared to running individual models sequentially.

The challenge of handling separate and conflicting datasets without requiring unified annotation was addressed, offering significant value for future research and real-world applications.

The proposed approach, based on YOLO, allows efficient sharing of visual features across different tasks. Hard parameter sharing and Representation Similarity Analysis (RSA) were employed to optimize task-specific performance while maintaining high computational efficiency.

Extensive experiments were conducted on both proprietary production-scale data and open-source datasets (PASCAL VOC and Objects365). The findings highlighted the model's superior performance and versatility. 

Experiments with the zero-shot approach showed that using OVD in a similar scenario across multiple datasets ($>2$) does not provide an advantage. In key metrics, especially those important in real-world applications such as speed and, most importantly, accuracy, OVD performs worse. This gap is explained by the need for a large amount of data for training, including target data, and the tendency to overfit on base classes. The only real advantage of the zero-shot approach in the current scenario is its speed and flexibility in adding potentially new and infinite categories, which can be useful in developing and testing class schemes. In practice, \ModelName\ can be viewed as an intermediate solution between traditional detectors and OVD, taking the best of both.

Furthermore, \ModelName\ is designed to be easily expandable, supporting additional tasks beyond object detection, such as attribute recognition and embedding calculation. 

To support the research community and practitioners, the framework—including algorithms' implementation, all necessary code, and the model trained on open-source data—has been made publicly available. These resources aim to facilitate further research and practical applications, providing a foundation for significant improvements and innovations in the future.

{\small
\bibliographystyle{ieee_fullname}
\bibliography{egbib}
}

\end{document}